%% file: main.tex
\documentclass[runningheads]{llncs}
\usepackage{graphicx}
\usepackage{amsmath}
\usepackage{subcaption}
\usepackage{arydshln}
\usepackage{hyperref}
\usepackage{amsfonts}
\usepackage{cleveref}
\usepackage{todonotes}
\usepackage{bbm}
\usepackage{algorithm}
\usepackage[noend]{algorithmic}
\usepackage{pgfplots}
\usepackage{placeins}
\usepackage[misc]{ifsym}

\newcommand{\q}{^q}
\newcommand{\algcmt}[1]{\{$\triangleright$ #1\}}
\newcommand{\commentout}[1]{}

\begin{document}
\sloppy
\title{Functional Latent Dynamics for Irregularly Sampled Time Series Forecasting}
%
%
\author{Christian Klötergens(\Letter)\inst{1}\inst{2} \and Vijaya Krishna Yalavarthi\inst{1} \and Maximilian Stubbemann\inst{1}\inst{2} \and Lars Schmidt-Thieme\inst{1}\inst{2}}

\tocauthor{Christian Klötergens, Vijaya Krishna Yalavarthi, Maximilian Stubbemann, Lars Schmidt-Thieme}
\toctitle{Functional Latent Dynamics for Irregularly Sampled Time Series Forecasting}

\institute{ISMLL, University of Hildesheim, Germany \email{\{kloetergens, yalavarthi, stubbemann, schmidt-thieme\}@ismll.de}
\and 
VWFS Data Analytics Research Center}
\authorrunning{Klötergens, et al.}
%
%
\maketitle              

\setlength{\tabcolsep}{5pt}
\renewcommand{\arraystretch}{1.3}

\input{content/00_abstract.tex}
\input{content/01_intro.tex}
\input{content/02_Problem-Formulation-Vijaya-2-lst.tex}
\input{content/03_background.tex}
\input{content/04_Method.tex}
\input{content/05_Toyexp.tex}

\input{content/06_a_Benchdetails.tex}

\input{content/06_b_benchresults.tex}
\input{content/06_c_efficency.tex}
\FloatBarrier
\input{content/07_Concl}
%
%
%
%
\bibliographystyle{splncs04}
\bibliography{references}
\end{document}

%% file: content/00_abstract.tex
\begin{abstract}
    Irregularly sampled time series with missing values are often observed
    in multiple real-world applications such as healthcare, climate and astronomy. They pose a significant challenge to standard deep 
    learning models that operate only on fully observed and regularly sampled
    time series.
    In order to capture the continuous dynamics of the irregular time series,
    many models rely on solving an Ordinary Differential Equation (ODE)
    in the hidden state. These ODE-based models tend to perform slow and require
    large memory due to sequential operations and a complex ODE solver.
    As an alternative to complex ODE-based models, 
    we propose a family of models called Functional Latent Dynamics (FLD).
    Instead of solving the ODE, we use simple curves which exist at all time points to 
    specify the continuous latent state in the model.
    The coefficients of these curves are learned only from the 
    observed values in the time series ignoring the missing values.
    Through extensive experiments, we demonstrate that
    FLD achieves better performance compared to the best ODE-based model
    while reducing the runtime and memory overhead. Specifically,
    FLD requires an order of magnitude less time to infer the forecasts
    compared to the best performing forecasting model.
    
\keywords{Irregularly Sampled Time Series \and Missing Values \and Forecasting}
\end{abstract}

%% file: content/01_intro.tex
\section{Introduction}
\begin{figure}
    \begin{minipage}{0.48\textwidth}
        \centering
        \begin{subfigure}{\linewidth}
            \includegraphics[width=\linewidth]{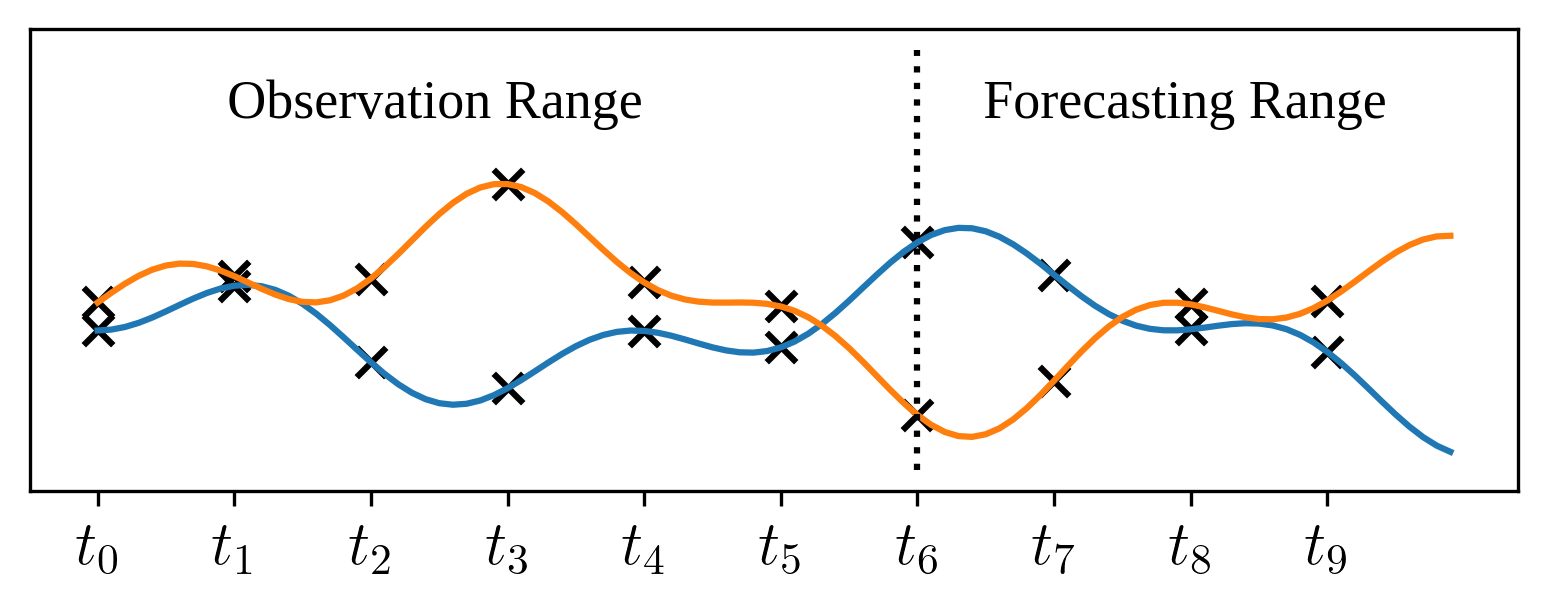}
            \caption{Regular Time Series}
            \label{fig:reg}
        \end{subfigure}
    \end{minipage}
    \hfill
    \begin{minipage}{0.48\textwidth}
        \centering
        \begin{subfigure}{\linewidth}
            \includegraphics[width=\linewidth]{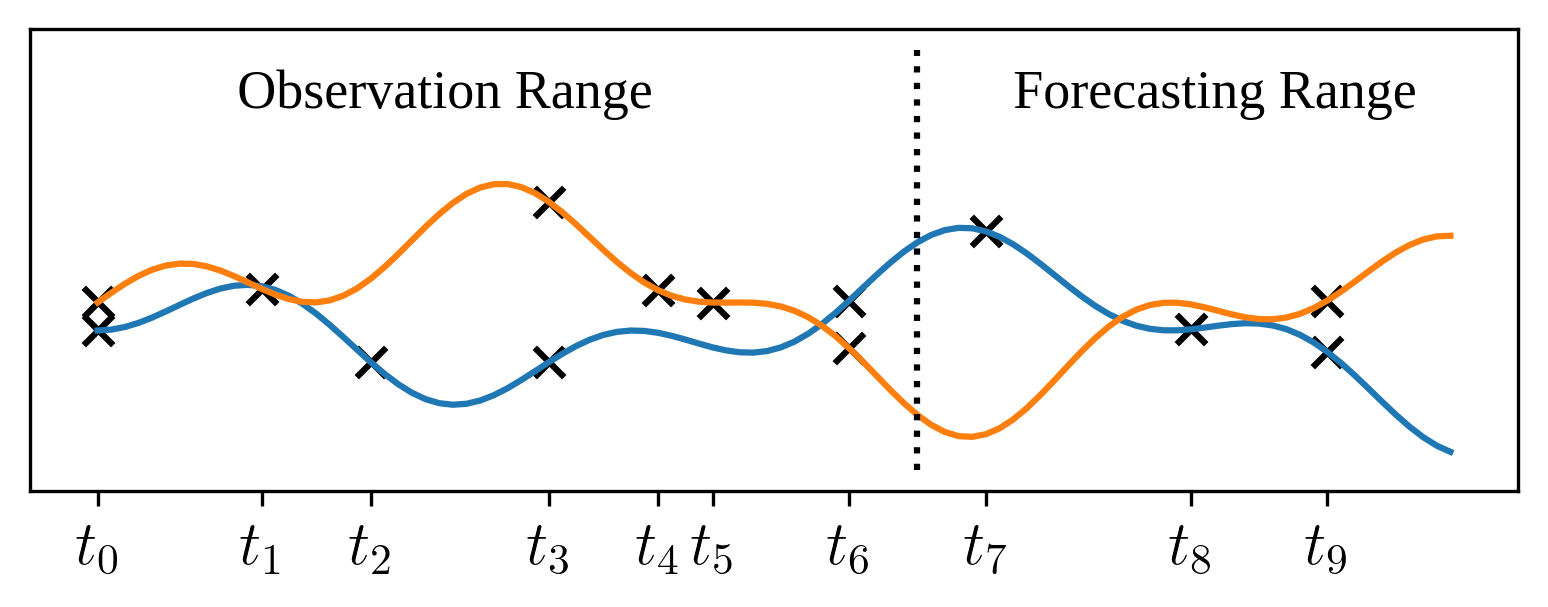}
            \caption{IMTS}
            \label{fig:imts}
        \end{subfigure}
    \end{minipage}
    \caption{Example for regularly and irregularly sampled Time Series with two channels. The observations and forecasting targets are marked as black crosses.}
    \label{fig:misv}
\end{figure}

Time series forecasting plays a pivotal role in numerous fields, ranging from
finance and economics to environmental science and healthcare. A time series is
considered multivariate if multiple variables, also known as channels, are
observed. In the realm of time series forecasting, most of the
literature considers \emph{regular time series}, where the
time difference between the observed points is equal, and no observations are
missing. However, in real-world application such as in healthcare domains,
different channels are often independently and irregularly observed leading to an extremely
sparse multivariate time series when they are aligned. We refer to these time series as 
Irregularly Sampled Multivariate Time Series with missing values (IMTS).
The forecasting task of regular multivariate time series and
IMTS is illustrated in Figure~\ref{fig:misv}.

Forecasting of IMTS is not well-covered in the literature compared to forecasting
of regular time series. 
Machine learning models that are designed for forecasting regular
multivariate time series often rely on the relative position of the observation in the series 
rather than
the absolute time, and cannot accommodate missing values. 
Applying these models to IMTS forecasting is not trivial. More specific, models need to implement strategies to handle varying observation distances and missing values.  
The standard method of handling missing values is imputation. However, this approach is 
usually suboptimal as absence of data itself carries information,
which is discarded by imputation.
Additionally, imputation errors accumulate and heavily affect the final forecasting task. Therefore, IMTS models must 
incorporate a more advanced method to handle
missing values and directly take observation times into account.



Ordinary Differential Equation (ODE)-based
models~\cite{chenneuralODE,de2019gru,bilovs2021neural,scholz2023latent,schirmer2022cru}
have been widely studied for this task. 
These models capture underlying dynamics of continuous time, making them
well-suited for IMTS forecasting where the time intervals between observations vary.
However, ODE-based models cannot directly handle the 
missing values, a prevalent occurrence in various application
scenarios. Furthermore, they are inefficient in terms of run time as they operate
in sequential manner similar to recurrent neural networks (RNNs). 

In this work, we propose a novel family of models
called \textbf{F}unctional \textbf{L}atent
\textbf{D}ynamics (FLD).
The hidden states of FLD are governed by a 
function whose coefficients are derived from the observed time series.
The hidden state function can be any curve such as a polynomial or sine function.
As the hidden state function accepts continuous time points as inputs, 
it can be evaluated at any desired time.
Our encoder considers only observed values in the time series and
ignores the missing values to parameterize the hidden state function. 
Finally, a dense fully connected deep neural network is applied to the
hidden state to obtain the forecasts.

Our approach is capable of utilizing any type of parameterized, differentiable 
function and can thus be adapted to various forecasting scenarios. 

FLD serves as an alternative to ODE-based models and
can handle both missing values and irregular sampling.
By employing simple curve functions to model hidden state dynamics,
we demonstrate that the forecasting accuracy of FLD
is significantly better than ODE based models and 
competitive with the state-of-the-art IMTS forecasting models
on $4$ real-world IMTS datasets.
Additional studies on computational efficiency show that
FLD significantly outperforms competing
models in terms of inference time.
Our contributions are as follows.

\begin{itemize}

    \item We propose Functional Latent Dynamics (FLD), a novel method for IMTS forecasting. FLD captures latent
    dynamics in a continuous fashion with parameterized curve functions.
    
    \item We propose an approach to incorporate the well-established attention
    mechanism to learn the coefficients of our curve functions that encode the IMTS.
    
    
    \item We provide a Proof-of-Concept on a simple toy dataset that is generated with the Goodwin oscillator model~\cite{goodwin1965oscillatory}, an ODE designed to model enzyme synthesis.    
    
    \item We conduct extensive experiments on established benchmark tasks. Our
    results indicate that FLD outperforms state-of-the-art competitors by an order of magnitude in terms of inference time while providing competitive forecasting
    accuracy.
\end{itemize}

Our code is publicly available on an anonymous Git repository:
\url{https://github.com/kloetergensc/Functional-Latent_Dynamics} 

%% file: content/02_Problem-Formulation-Vijaya-2-lst.tex
\newcommand{\obs}{^\text{obs}}
\newcommand{\R}{\mathbb{R}}
\newcommand*{\N}{\mathbb{N}}
\newcommand{\nan}{\text{NaN}}
\newcommand{\M}{\mathcal{M}}


\section{Problem Formulation}
%

An \emph{Irregularly sampled multivariate time series (IMTS)}
is a sequence $x := \left(\left(t_1, v_1\right),
\ldots, \left(t_N, v_N\right)\right)$
of $N$ many pairs where each pair
consists of an observation time point $t_n \in \R$
and observation event $v_n \in X^C:= (\R \, \cup \, \{\text{NaN}\})^C$ 
made at $t_n$; $C \in \N$ is the number of
channels, $v_{n,c} \neq \nan$ represents an \emph{observed value} and $ v_{n,c} =
\nan$ represents a \emph{missing value}.
An \emph{IMTS forecasting query} is a sequence
$t\q := (t\q_1, \ldots, t\q_K)$
of time points for which observation values are
sought (where $\displaystyle \min_{k=1:K} \, t\q_k > \max_{n=1:N} \, t_n$).
Any sequence $y := (y_1, \ldots, y_K)$ of same length
with values in $X^C$ we call
an \emph{IMTS forecasting answer}.
To measure the difference between
  the ground truth forecasting answer $y$ (possibly with missing values) and
  the predicted forecasting answer $\hat{y}$ (without missing values),
a scalar loss function $\ell : \R\times\R  \to \R$
such as squared error 
is averaged over all query time points and non-missing observations:
\begin{align*}
    \ell(y, \hat{y}) := 
    \frac{1}{\sum_{k=1}^{K} N_k}~\sum_{k=1}^K\sum_{\substack{c=1 \\ y_{k,c}\ne \nan}}^C \ell(y_{k,c} , \hat{y}_{k,c}),
\end{align*}
where $N_k= |\{c \in [C] \mid y_{k,c} \neq \nan \}|$ denotes the amount of
non-missing values of the forecasting answer $y_k$ at time point $t^q_k$.

An \emph{IMTS forecasting dataset} consists of $M$ many triples
$(x_m,t_m\q,y_m)$ (called instances) consisting of
  a past IMTS $x_m$,
  a query $t_m\q$ of future time points and
  the ground truth observation values $y_m$ for those time points,
drawn from an unknown distribution $\rho$.
The length $N$ of the past and
  the number $K$ of queries 
  will vary across instances in general,
while the number $C$ of channels is the same for all instances.

The \emph{IMTS forecasting problem} then is,
 given
   such a dataset $D := ( (x_1, t\q_1, y_1), \ldots, (x_M, t\q_M, y_M))$ and
   a loss function $\ell$,
find a model
  $\M: (X^C)^* \times \R^* \to (\R^C)^*$,
where $^*$ denotes finite sequences,
such that its expected forecasting loss is
minimal:
\begin{align*}
 \mathcal{L}(\M; \rho) := \mathop\mathbb{E}_{(x, t\q, y)\sim \rho} 
 [\ell(y, \M(x,t\q))]
\end{align*}





%% file: content/03_background.tex
\section{Background}
\label{sec:bkgd}

ODE-based models~\cite{chenneuralODE,de2019gru,bilovs2021neural,schirmer2022cru,scholz2023latent}
are a family of continuous-time models wherein
the hidden state $z(\tau)$ is the solution of an
initial value problem in Ordinary Differential Equations (ODEs):
\begin{align}
    \frac{dz(\tau)}{d\tau} = f(\tau, z(\tau))
    \quad \text{where} \quad z(\tau_0) = z_0
\end{align}
Here, $\tau$ can both reference to observation time points $t$ and query time points $t^q$.
$f$ is a trainable neural network that governs the dynamics 
of the hidden state.
The hidden state $z(\tau)$ is defined
and can be evaluated at any desired time-point.
Hence, they are a natural fit to model IMTS, where observation times are continuous.
However, a numerical ODE solver is required to infer the hidden
state:
\begin{align}
    z_0, \ldots, z_N := \text{ODESolve}\left(f, 
    z_0, \left(\tau_0,\ldots,\tau_N\right)\right) \label{eq:odesolve}
\end{align}
Here, $z_n$ is the hidden state for $\tau_n$ and $z_0$ is the 
initial value.

GRU-ODE-Bayes~\cite{de2019gru} integrates a continuous version of 
Gated Recurrent Units (GRU) into the neural ODE architecture and updates $z(t)$ 
with Bayesian inference. 

LinODENet~\cite{scholz2023latent} replaces the neural ODE with a linear ODE, 
in which the ODE solutions are computed by a linear layer. Using a linear ODE enables 
the model to omit the ODE solver.
For updates at observations, LinODENet model incorporates Kalman filtering to
ensure the self-consistency property, where the state of the model only changes
when the observation deviates from the model prediction. 

Continuous Recurrent Units (CRU)~\cite{schirmer2022cru} replace the ODE with a
Stochastic Differential Equation (SDE). 
Using an SDE has the benefit that the change of latent state over any time frame can
be computed in closed form with continuous-discrete Kalman filtering. 

Related to neural ODE, Neural Flows~\cite{bilovs2021neural} apply invertible networks to
directly model the solution curves of ODEs, rendering the ODE solver
obsolete.

While ODE-based models have the advantage of learning from
continuous time observations, they require a complex numerical
ODE solver which is slow~\cite{bilovs2021neural}. Additionally,
they process the observations of an IMTS sequentially
worsening the run time and also increase the memory requirements.
Furthermore, ODE-based models cannot directly handle missing values. 
Typically, they require missing value indicators which act as additional
input channels in the series complicating the learning process. 

Substantially different from neural ODEs, GraFITi~\cite{grafiti} encodes 
time series as graphs and solves the forecasting
problem using graph neural networks. The model showed superior forecasting accuracy on
the established benchmark datasets, while having significantly faster inference than ODE-based models.  

%% file: content/04_Method.tex
\newcommand{\enc}{\text{FLD-Encoder}}
\section{Functional Latent Dynamics}
\label{sec:FLD}

\label{sec:chs}

\begin{figure}[t]
    \centering
    \small
    \input{content/fig/overall_arch}
    \caption{Example of FLD with sine functions as a
    3-dimensional hidden state. The parameters $\theta$ of the hidden 
    state function $g(\cdot; \theta)$  are inferred by aggregating 
    the observations (red/blue dots) with the attention-based FLD-Encoder. 
    The hidden state at the query times is acquired 
    by following $g(t^q; \theta)$ and decoded by a neural network ($\text{NN}^\text{out}$).}
	\label{fig:network}
\end{figure}
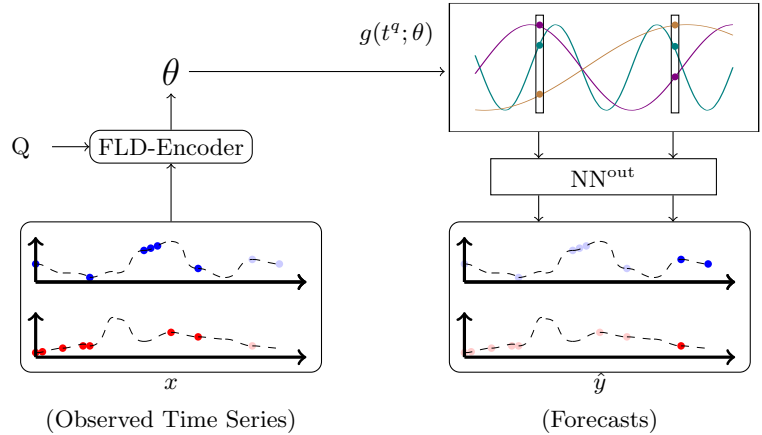

We introduce a family of models called Functional Latent Dynamics
as an alternative to ODE-based models.
Here, we use simple curves to specify the hidden state.
Specifically, we replace ODESolve in \cref{eq:odesolve}
with curves such as polynomial or sine functions.
The latent state $z_n$ is given as:
\begin{align*}
    z_n := g(\tau_n; \theta)
\end{align*}
where $g$ is a curve with coefficients $\theta$.

Inferring the hidden state at any time point with
a function is computationally efficient if 
that function is simple and does not depend on other time points.
Large portion of the literature applies sequential models
such as RNNs to learn the inductive bias from the causal nature of
time series. However, they are slow, as they have
to operate sequentially. Alternatively,
recent transformer based works in similar
domains~\cite{patchtst,vit}
show that we can achieve state-of-the-art performance even without
applying the sequential model. Hence, in this work, we use simple curves
such as polynomial (linear (FLD-L) in \cref{eq:FLD-linear},
quadratic (FLD-Q) in \cref{eq:FLD-quad}) or sine (FLD-S) functions (in \cref{eq:FLD-sin}).

\begin{align}
    g^\text{lin}(t;\theta) &:= \theta_1 t + \theta_2 && \theta = (\theta_1, \theta_2) \in \mathbb{R}^{2 \times L} \label{eq:FLD-linear} \\
    g^\text{quad}(t;\theta) &:= \theta_1 t^2 + \theta_2 t + \theta_3 && \theta = (\theta_1, \theta_2, \theta_3) \in \mathbb{R}^{3 \times L} \label{eq:FLD-quad} \\
    g^\text{sin}(t;\theta) &:= \theta_1\sin(\theta_2 + \theta_3t) + \theta_4 && \theta = (\theta_1, \theta_2, \theta_3, \theta_4) \in \mathbb{R}^{4 \times L}\label{eq:FLD-sin}
\end{align}

\commentout{
The problem with sine and quadratic functions
is that they are not injective meaning, they can map
different $t$ to same value which is not desired.
Inspired from general positional embedding in
Transformers~\cite{vaswani2017attention}
we use a $L$ dimensional $\theta$ 
($\theta \in \R^{R\times L}$, $|\theta| = R$) so that
the function becomes near injective for finite many time points.
We demonstrate the dynamics of the functions in Figure~\ref{fig:curves}.
}

Here, \emph{sin} is applied coordinate-wise. Once we have computed the latent state $z$, we apply
a multilayer feedforward neural network ($\text{NN}^\text{out}$) to compute
$\hat{y}$ via
\begin{align*}
    \hat{y}_k = \text{NN}^\text{out}(z_k).
\end{align*}

\begin{algorithm}[t]
\caption{Functional Linear Dynamics}
\label{alg:FLD}
\begin{algorithmic}[1]
    \REQUIRE $\text{Observed IMTS } x, \text{ Query time points } 
    t\q, \text{ latent function }g$ 
    \STATE $\theta \gets \enc(x)$ \hfill \algcmt{Compute the function coefficients}
    \FOR{$k = 1, \ldots ,K$}
    \STATE $z_k \gets g(t\q_k, \theta)$ \hfill \algcmt{Compute the latent state}
    \STATE $\hat{y}_k\gets \text{NN}^\text{out}(z_k)$ \hfill \algcmt{Make the prediction}
    \ENDFOR
    \RETURN $(\hat{y}_k)_{k=1:K}$
\end{algorithmic}
\end{algorithm}

\section{Inferring Coefficients}\label{sec:enc}

Values of $\theta$ are computed from the observed time series $X$ using the FLD-Encoder.
First, we convert $X$ into $C$ many tuples
 $x^{(1)}, \ldots, x^{(C)}$ where $x^{(c)}=(t(c),v(c))$.
Here, $t^{(c)}=(t_1^{(c)}, \ldots, t^{(c)}_{N_c})$ and $v^{(c)}=(v_1^{(c)}, \ldots, v^{(c)}_{N_c})$
represent the observation time points
and values in channel $c$, respectively, i.e., the time points with no missing
values in channel $c$ and the corresponding values.
We pass all the tuples $x^{(c)}$ to a multi-head attention based encoder.
We begin with time embeddings.

\paragraph{Continuous time embeddings.}
Our attention-based FLD-Encoder consists of $H$ many heads and for each head $h$,
we provide a $D$ dimensional embedding 
$\phi^h: \R \to \R^D$ of time points:
\begin{align}
    \phi^{h}_d(t):= \begin{cases}
        a_{dh} t + b_{dh} \quad & \text{if } d=1 \\
        \sin(a_{dh} t + b_{dh}) \quad & \text{if } 1 < d \le D 
    \end{cases}
\end{align}
Here, $a_{dh}$ and $b_{dh}$ are trainable parameters.
This embedding helps to learn periodic terms from the sinusoidal
embeddings and non-periodic
terms from the linear embedding~\cite{shukla2021mTAN}.

\paragraph*{Multi-head attention encoder.}
In the following, $\text{Q}^h\in \R^{R\times D}$ is
a matrix of trainable parameters where $\text{Q}^h_r$
provides vector representation to $\theta_r$, $R = |\theta|$.
$\text{K}^{h,c} := \phi_h(t^{(c)})$ is the continuous embedding of time points in $t^{(c)}$,
and $\text{V}^c = v^{(c)}$. $\text{FF} : \mathbb{R}^{HC} \to \mathbb{R}^L$ is a single feed forward layer. 
Note that similar to scaled dot-product attention
in~\cite{vaswani2017attention}, softmax is applied row wise.

Presence of missing values in the data makes it challenging
to apply multi-head attention directly. Hence, we modify it
as follows:
\begin{align*}
    \theta & := \text{FF}(\hat{\theta}) && \in \R^{R\times L} \\
    \hat{\theta} &:= [\hat{\theta}^{1,1}, \ldots, \hat{\theta}^{1,C}, 
    \ldots, \hat{\theta}^{H,1}, \ldots, \hat{\theta}^{H,C}  ] && \in \R^{R \times HC}\\ 
    \hat{\theta}^{h,c} &:= A^{h,c} \text{V}^{c}  && \in \R^{R\times 1}\\
    A^{h,c} & := \text{softmax}\left(\text{Q}^h\left(\text{K}^{h,c}\right)^T/\sqrt{D} \right) && \in \R^{R \times |N_c|}
\end{align*}

A forward pass of IMTS forecasting using the proposed model is
presented in Algorithm~\ref{alg:FLD}.

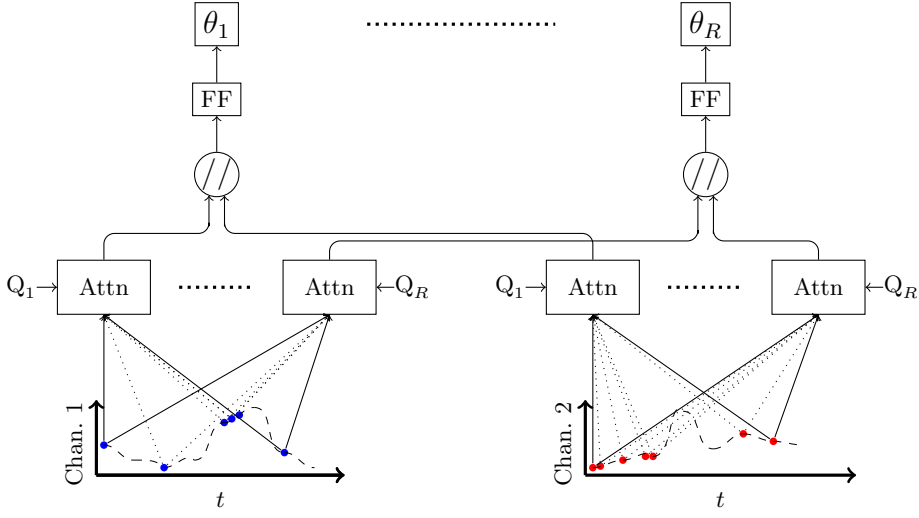
\begin{figure}[t]
	\centering
	\small
	\input{content/fig/encoder}
	\caption{FLD-Encoder infers coefficients $\theta$ to model the hidden dynamics of an IMTS. The channel observations are aggregated 
     with attention (Attn), concatenated ($//$) and combined with a feed forward layer (FF).}
\end{figure}

\paragraph*{Delineating from mTAN Encoder.}
Our encoder shares some features with the mTAN encoder~\cite{shukla2021mTAN}.
The mTAN encoder is used to convert an IMTS into a fully observed 
regularly sampled time series in the latent space.
Instead, the goal of our encoder is to compute the coefficients $\theta$
instead of converting to another time series. 
Hence, our attention query is a 
trainable matrix instead of embedded reference time points.

%% file: content/fig/overall_arch.tex
\begin{tikzpicture}

	\node[draw, rounded corners, minimum height=2cm, minimum width=4cm, label={[yshift=-3cm]\begin{tabular}{c}
			$x$ \\
			(Observed Time Series)
		\end{tabular}}] at (1.8,-0.2) (inp_ts) {};
	\begin{scope}[yshift=0cm, xshift=0cm, xscale=0.9,yscale=0.6]
		\node (origin) at (0,0) {};
		\node (x) at (4,0)	{};
		\node (y) at (0,1)	{};
		\node[fill=blue, circle, inner sep=0, minimum size=1mm] (1x1) at (0,0.4) {};
		\node[fill=none, circle, inner sep=0, minimum size=1mm] (1x2) at (0.4, 0.2) {};
		\node[fill=blue, circle, inner sep=0, minimum size=1mm] (1x3) at (0.8, 0.1) {};
		\node[fill=none, circle, inner sep=0, minimum size=1mm] (1x4) at (1.2, 0.2) {};
		\node[fill=blue, circle, inner sep=0, minimum size=1mm] (1x5) at (1.6, 0.7) {};
		\node[fill=blue, circle, inner sep=0, minimum size=1mm] (151) at (1.7, 0.75) {};
		\node[fill=blue, circle, inner sep=0, minimum size=1mm] (152) at (1.8, 0.8) {};
		\node[fill=none, circle, inner sep=0, minimum size=1mm] (1x6) at (2.0, 0.9) {};
		\node[fill=blue, circle, inner sep=0, minimum size=1mm] (1x7) at (2.4, 0.3) {};
		\node[fill=none, circle, inner sep=0, minimum size=1mm] (1x8) at (2.8, 0.1) {};
		\node[fill=blue!20, circle, inner sep=0, minimum size=1mm] (1x9) at (3.2, 0.5) {};
		\node[fill=blue!20, circle, inner sep=0, minimum size=1mm] (1x10) at (3.6, 0.4) {};
		\path[-, dash pattern={on 3pt off 3pt}, line width=0.2pt] (1x1.center) edge[out=0, in=180] (1x2.center);
		\path[-, dash pattern={on 3pt off 3pt}, line width=0.2pt] (1x2.center) edge[out=0. in=180] (1x3.center);
		\path[-, dash pattern={on 3pt off 3pt}, line width=0.2pt] (1x3.center) edge[out=0. in=180] (1x4.center);
		\path[-, dash pattern={on 3pt off 3pt}, line width=0.2pt] (1x4.center) edge[out=0, in=180] (1x5.center);
		\path[-, dash pattern={on 3pt off 3pt}, line width=0.2pt] (1x5.center) edge[out=0, in=180] (1x6.center);
		\path[-, dash pattern={on 3pt off 3pt}, line width=0.2pt] (1x6.center) edge[out=0, in=180] (1x7.center);
		\path[-, dash pattern={on 3pt off 3pt}, line width=0.2pt] (1x7.center) edge[out=0, in=180] (1x8.center);
		\path[-, dash pattern={on 3pt off 3pt}, line width=0.2pt] (1x8.center) edge[out=0, in=180] (1x9.center);
		\path[-, dash pattern={on 3pt off 3pt}, line width=0.2pt] (1x9.center) edge[out=0, in=180] (1x10.center);
		\path[->, very thick] (origin.center) edge (x.center);
		\path[->, very thick] (origin.center) edge  (x.center);
		\path[->, very thick] (origin.center) edge (y.center);
	\end{scope}

	\begin{scope}[yshift=-1cm, xshift=-0cm, xscale=0.9,yscale=0.6]
	\node (origin) at (0,0) {};
	\node (x) at (4,0)	{};
	\node (y) at (0,1)	{};
	\node[fill=red, circle, inner sep=0, minimum size=1mm] (x1) at (0,0.1) {};
	\node[fill=red, circle, inner sep=0, minimum size=1mm] (10) at (0.1,0.12) {};
	\node[fill=red, circle, inner sep=0, minimum size=1mm] (x2) at (0.4, 0.2) {};
	\node[fill=red, circle, inner sep=0, minimum size=1mm] (x20) at (0.7,0.25) {};
	\node[fill=red, circle, inner sep=0, minimum size=1mm] (x3) at (0.8, 0.25) {};
	\node[fill=none, circle, inner sep=0, minimum size=1mm] (x4) at (1.2, 0.85) {};
	\node[fill=none, circle, inner sep=0, minimum size=1mm] (x5) at (1.6, 0.35) {};
	\node[fill=red, circle, inner sep=0, minimum size=1mm] (x6) at (2.0, 0.55) {};
	\node[fill=red, circle, inner sep=0, minimum size=1mm] (x7) at (2.4, 0.45) {};
	\node[fill=none, circle, inner sep=0, minimum size=1mm] (x8) at (2.8, 0.4) {};
	\node[fill=red!20, circle, inner sep=0, minimum size=1mm] (x9) at (3.2, 0.25) {};
	\node[fill=none, circle, inner sep=0, minimum size=1mm] (x10) at (3.6, 0.2) {};
	\path[-, dash pattern={on 3pt off 3pt}, line width=0.2pt] (x1.center) edge[out=0, in=180] (x2.center);
	\path[-, dash pattern={on 3pt off 3pt}, line width=0.2pt] (x2.center) edge[out=0. in=180] (x3.center);
	\path[-, dash pattern={on 3pt off 3pt}, line width=0.2pt] (x3.center) edge[out=0. in=180] (x4.center);
	\path[-, dash pattern={on 3pt off 3pt}, line width=0.2pt] (x4.center) edge[out=0, in=180] (x5.center);
	\path[-, dash pattern={on 3pt off 3pt}, line width=0.2pt] (x5.center) edge[out=0, in=180] (x6.center);
	\path[-, dash pattern={on 3pt off 3pt}, line width=0.2pt] (x6.center) edge[out=0, in=180] (x7.center);
	\path[-, dash pattern={on 3pt off 3pt}, line width=0.2pt] (x7.center) edge[out=0, in=180] (x8.center);
	\path[-, dash pattern={on 3pt off 3pt}, line width=0.2pt] (x8.center) edge[out=0, in=180] (x9.center);
	\path[-, dash pattern={on 3pt off 3pt}, line width=0.2pt] (x9.center) edge[out=0, in=180] (x10.center);
	\path[->, very thick] (origin.center) edge (x.center);
	\path[->, very thick] (origin.center) edge (y.center);
\end{scope}

	\node[draw, above of=inp_ts, node distance=2cm, rounded corners] (enc) {FLD-Encoder};
	\node[draw=none, above of = enc, node distance=1cm] (theta) {\Large $\theta$};
	\node[draw=none, left of=enc, node distance=2.0cm] (Q) {\begin{tabular}{c} Q
	\end{tabular}};
	\node[draw=none, right of= theta, node distance=5.75cm] (curves) {};
	\node[draw, below of= curves, node distance=1.4cm, minimum width=3cm] (nnout) {$\text{NN}^{\text{out}}$};
	\draw[->] (Q) -- (enc);
	\draw[->] (inp_ts) -- (enc);
	\draw[->] (enc) -- (theta);
	\draw[->] (theta) -- ([xshift=3.4cm]theta.east);
	\draw[->] ([yshift=0.35cm]nnout.164) -- (nnout.164);
	\draw[->] ([yshift=0.35cm]nnout.15) -- (nnout.15);
	\draw[<-] ([yshift=-0.35cm]nnout.196) -- (nnout.196);
	\draw[<-] ([yshift=-0.35cm]nnout.345) -- (nnout.345);
	\node () at ([xshift=3cm, yshift=0.5cm]theta) {$g(t\q;\theta)$};
\begin{scope}[xshift=5.5cm, yshift=2cm,xscale = 0.6, yscale=0.3]
	\begin{axis}[
		minor tick num=0,
		ticks =none,
		ymax=1.5,
		ymin=-1.5,
		]
		\addplot[smooth,teal,thick,mark=none,
		domain=-1:1,samples=40]
		{cos(deg(8*x + 3))};
		\addplot[smooth,violet,thick,mark=none,
		domain=-1:1,samples=40]
		{cos(deg(4*x - 4))};
		\addplot[smooth,brown,thick,mark=none,
		domain=-1:1,samples=40]
		{cos(deg(2*x + 5))};
	\end{axis}
	\node[draw, minimum height=1.3cm, minimum width=1mm, inner sep=0] at (5,3) {};
	\node[draw, minimum height=1.3cm, minimum width=1mm, inner sep=0] at (2,3) {};
	\node at (5,2.35) {\Huge \color{violet} $\cdot$};
	\node at (5,3.7) {\Huge \color{teal} $\cdot$};
	\node at (5,4.65) {\Huge \color{brown} $\cdot$};

	\node at (2,1.58) {\Huge \color{brown} $\cdot$};
	\node at (2,3.75) {\Huge \color{teal} $\cdot$};
	\node at (2,4.65) {\Huge \color{violet} $\cdot$};

\end{scope}
\node[draw, rounded corners, minimum height=2cm, minimum width=4cm, label={[yshift=-3cm]\begin{tabular}{c}
		$\hat{y}$ \\
		(Forecasts)
\end{tabular}}] at (7.5,-0.2) (inp_ts) {};
	\begin{scope}[yshift=0cm, xshift=5.7cm, xscale=0.9,yscale=0.6]
	\node (origin) at (0,0) {};
	\node (x) at (4,0)	{};
	\node (y) at (0,1)	{};
	\node[fill=blue!20, circle, inner sep=0, minimum size=1mm] (1x1) at (0,0.4) {};
	\node[fill=none, circle, inner sep=0, minimum size=1mm] (1x2) at (0.4, 0.2) {};
	\node[fill=blue!20, circle, inner sep=0, minimum size=1mm] (1x3) at (0.8, 0.1) {};
	\node[fill=none, circle, inner sep=0, minimum size=1mm] (1x4) at (1.2, 0.2) {};
	\node[fill=blue!20, circle, inner sep=0, minimum size=1mm] (1x5) at (1.6, 0.7) {};
	\node[fill=blue!20, circle, inner sep=0, minimum size=1mm] (151) at (1.7, 0.75) {};
	\node[fill=blue!20, circle, inner sep=0, minimum size=1mm] (152) at (1.8, 0.8) {};
	\node[fill=none, circle, inner sep=0, minimum size=1mm] (1x6) at (2.0, 0.9) {};
	\node[fill=blue!20, circle, inner sep=0, minimum size=1mm] (1x7) at (2.4, 0.3) {};
	\node[fill=none, circle, inner sep=0, minimum size=1mm] (1x8) at (2.8, 0.1) {};
	\node[fill=blue, circle, inner sep=0, minimum size=1mm] (1x9) at (3.2, 0.5) {};
	\node[fill=blue, circle, inner sep=0, minimum size=1mm] (1x10) at (3.6, 0.4) {};
	\path[-, dash pattern={on 3pt off 3pt}, line width=0.2pt] (1x1.center) edge[out=0, in=180] (1x2.center);
	\path[-, dash pattern={on 3pt off 3pt}, line width=0.2pt] (1x2.center) edge[out=0. in=180] (1x3.center);
	\path[-, dash pattern={on 3pt off 3pt}, line width=0.2pt] (1x3.center) edge[out=0. in=180] (1x4.center);
	\path[-, dash pattern={on 3pt off 3pt}, line width=0.2pt] (1x4.center) edge[out=0, in=180] (1x5.center);
	\path[-, dash pattern={on 3pt off 3pt}, line width=0.2pt] (1x5.center) edge[out=0, in=180] (1x6.center);
	\path[-, dash pattern={on 3pt off 3pt}, line width=0.2pt] (1x6.center) edge[out=0, in=180] (1x7.center);
	\path[-, dash pattern={on 3pt off 3pt}, line width=0.2pt] (1x7.center) edge[out=0, in=180] (1x8.center);
	\path[-, dash pattern={on 3pt off 3pt}, line width=0.2pt] (1x8.center) edge[out=0, in=180] (1x9.center);
	\path[-, dash pattern={on 3pt off 3pt}, line width=0.2pt] (1x9.center) edge[out=0, in=180] (1x10.center);
	\path[->, very thick] (origin.center) edge (x.center);
	\path[->, very thick] (origin.center) edge  (x.center);
	\path[->, very thick] (origin.center) edge (y.center);
\end{scope}

\begin{scope}[yshift=-1cm, xshift=5.7cm, xscale=0.9,yscale=0.6]
	\node (origin) at (0,0) {};
	\node (x) at (4,0)	{};
	\node (y) at (0,1)	{};
	\node[fill=red!20, circle, inner sep=0, minimum size=1mm] (x1) at (0,0.1) {};
	\node[fill=red!20, circle, inner sep=0, minimum size=1mm] (10) at (0.1,0.12) {};
	\node[fill=red!20, circle, inner sep=0, minimum size=1mm] (x2) at (0.4, 0.2) {};
	\node[fill=red!20, circle, inner sep=0, minimum size=1mm] (x20) at (0.7,0.25) {};
	\node[fill=red!20, circle, inner sep=0, minimum size=1mm] (x3) at (0.8, 0.25) {};
	\node[fill=none, circle, inner sep=0, minimum size=1mm] (x4) at (1.2, 0.85) {};
	\node[fill=none, circle, inner sep=0, minimum size=1mm] (x5) at (1.6, 0.35) {};
	\node[fill=red!20, circle, inner sep=0, minimum size=1mm] (x6) at (2.0, 0.55) {};
	\node[fill=red!20, circle, inner sep=0, minimum size=1mm] (x7) at (2.4, 0.45) {};
	\node[fill=none, circle, inner sep=0, minimum size=1mm] (x8) at (2.8, 0.4) {};
	\node[fill=red, circle, inner sep=0, minimum size=1mm] (x9) at (3.2, 0.25) {};
	\node[fill=none, circle, inner sep=0, minimum size=1mm] (x10) at (3.6, 0.2) {};
	\path[-, dash pattern={on 3pt off 3pt}, line width=0.2pt] (x1.center) edge[out=0, in=180] (x2.center);
	\path[-, dash pattern={on 3pt off 3pt}, line width=0.2pt] (x2.center) edge[out=0. in=180] (x3.center);
	\path[-, dash pattern={on 3pt off 3pt}, line width=0.2pt] (x3.center) edge[out=0. in=180] (x4.center);
	\path[-, dash pattern={on 3pt off 3pt}, line width=0.2pt] (x4.center) edge[out=0, in=180] (x5.center);
	\path[-, dash pattern={on 3pt off 3pt}, line width=0.2pt] (x5.center) edge[out=0, in=180] (x6.center);
	\path[-, dash pattern={on 3pt off 3pt}, line width=0.2pt] (x6.center) edge[out=0, in=180] (x7.center);
	\path[-, dash pattern={on 3pt off 3pt}, line width=0.2pt] (x7.center) edge[out=0, in=180] (x8.center);
	\path[-, dash pattern={on 3pt off 3pt}, line width=0.2pt] (x8.center) edge[out=0, in=180] (x9.center);
	\path[-, dash pattern={on 3pt off 3pt}, line width=0.2pt] (x9.center) edge[out=0, in=180] (x10.center);
	\path[->, very thick] (origin.center) edge (x.center);
	\path[->, very thick] (origin.center) edge (y.center);
\end{scope}

\end{tikzpicture}

%% file: content/fig/encoder.tex
\begin{tikzpicture}
	\node[draw, minimum height=5mm, minimum width=5mm] (theta1) at (0,1) {\large $\theta_1$};
	\node[draw, minimum height=5mm, minimum width=5mm] (theta2) at (6.5,1) {\large $\theta_R$};
	\node[draw] (FF1) at (0,0) {$\text{FF}$};
	\node[draw] (FF2) at (6.5,0) {$\text{FF}$};
	\node[draw, circle, inner sep=0] (cat1) at (0,-1) {\large //};
	\node[draw, circle, inner sep=0] (cat2) at (6.5,-1) {\large //};
	\node[draw=none, inner sep=0] (q11) at (-2.6,-2.5) {$\text{Q}_1$};
	\node[draw] (atn11) at (-1.5,-2.5) {\begin{tabular}{c}
			Attn\end{tabular}};
	\node[draw=none, inner sep=0] (q21) at (2.6,-2.5) {$\text{Q}_R$};
	\node[draw] (atn21) at (1.5,-2.5) {\begin{tabular}{c}
			Attn\end{tabular}};
	\node[draw=none, inner sep=0] (q12) at (3.9,-2.5) {$\text{Q}_1$};
	\node[draw] (atn12) at (5,-2.5) {\begin{tabular}{c}
			Attn\end{tabular}};
	\node[draw=none, inner sep=0] (q22) at (9.1,-2.5) {$\text{Q}_R$};
	\node[draw] (atn22) at (8,-2.5) {\begin{tabular}{c}
			Attn\end{tabular}};
	\draw[->] (FF1) -- (theta1);
	\draw[->] (FF2) -- (theta2);
	\draw[->] (cat1) -- (FF1);
	\draw[->] (cat2) -- (FF2);
	\draw[->] (q11) -- (atn11);
	\draw[->] (q12) -- (atn12);
	\draw[->] (q21) -- (atn21);
	\draw[->] (q22) -- (atn22);
	\draw[dotted, very thick] (-0.5,-2.5) -- (0.5,-2.5);
	\draw[dotted, very thick] (6,-2.5) -- (7,-2.5);
	\draw[dotted, very thick] (2,1) -- (4.5,1);
	\draw[<-, rounded corners] (cat1.250) |- +(0,-5mm) -| (atn11.north);
	\draw[<-, rounded corners] (cat1.290) |- +(0,-5mm) -| (atn12.north);
		\draw[<-, rounded corners] (cat2.250) |- +(0,-6mm) -| (atn21.north);
	\draw[<-, rounded corners] (cat2.290) |- +(0,-6mm) -| (atn22.north);
	\begin{scope}[yshift=-5cm, xshift=5cm]
		\node (origin) at (-0.1,0) {};
		\node (x) at (3.5,0)	{};
		\node (y) at (-0.1,1)	{};
		\node[fill=red, circle, inner sep=0, minimum size=1mm] (x1) at (0,0.1) {};
		\node[fill=red, circle, inner sep=0, minimum size=1mm] (10) at (0.1,0.12) {};
		\node[fill=red, circle, inner sep=0, minimum size=1mm] (x2) at (0.4, 0.2) {};
		\node[fill=red, circle, inner sep=0, minimum size=1mm] (x20) at (0.7,0.25) {};
		\node[fill=red, circle, inner sep=0, minimum size=1mm] (x3) at (0.8, 0.25) {};
		\node[fill=none, circle, inner sep=0, minimum size=1mm] (x4) at (1.2, 0.85) {};
		\node[fill=none, circle, inner sep=0, minimum size=1mm] (x5) at (1.6, 0.35) {};
		\node[fill=red, circle, inner sep=0, minimum size=1mm] (x6) at (2.0, 0.55) {};
		\node[fill=red, circle, inner sep=0, minimum size=1mm] (x7) at (2.4, 0.45) {};
		\node[fill=none, circle, inner sep=0, minimum size=1mm] (x8) at (2.8, 0.4) {};
		\path[-, dash pattern={on 3pt off 3pt}, line width=0.2pt] (x1.center) edge[out=0, in=180] (x2.center);
		\path[-, dash pattern={on 3pt off 3pt}, line width=0.2pt] (x2.center) edge[out=0. in=180] (x3.center);
		\path[-, dash pattern={on 3pt off 3pt}, line width=0.2pt] (x3.center) edge[out=0. in=180] (x4.center);
		\path[-, dash pattern={on 3pt off 3pt}, line width=0.2pt] (x4.center) edge[out=0, in=180] (x5.center);
		\path[-, dash pattern={on 3pt off 3pt}, line width=0.2pt] (x5.center) edge[out=0, in=180] (x6.center);
		\path[-, dash pattern={on 3pt off 3pt}, line width=0.2pt] (x6.center) edge[out=0, in=180] (x7.center);
		\path[-, dash pattern={on 3pt off 3pt}, line width=0.2pt] (x7.center) edge[out=0, in=180] (x8.center);
		\path[->, very thick] (origin.center) edge node[below] [yshift=-1mm]{$t$} (x.center);
		\path[->, very thick] (origin.center) edge node[left, rotate=90, xshift=6mm, yshift=3mm] {Chan. 2} (y.center);
	\end{scope}

\draw[->] (x1) -- (atn12.south);
\draw[->, dotted] (10) -- (atn12.south);
\draw[->, dotted] (x2) -- (atn12.south);
\draw[->, dotted] (x20) -- (atn12.south);
\draw[->, dotted] (x3) -- (atn12.south);
\draw[->, dotted] (x6) -- (atn12.south);
\draw[->] (x7) -- (atn12.south);

\draw[->] (x1) -- (atn22.south);
\draw[->, dotted] (10) -- (atn22.south);
\draw[->, dotted] (x2) -- (atn22.south);
\draw[->, dotted] (x20) -- (atn22.south);
\draw[->, dotted] (x3) -- (atn22.south);
\draw[->, dotted] (x6) -- (atn22.south);
\draw[->] (x7) -- (atn22.south);

\begin{scope}[yshift=-5cm, xshift=-1.5cm]
	\node (origin) at (-0.1,0) {};
	\node (x) at (3.2,0)	{};
	\node (y) at (-0.1,1)	{};
	\node[fill=blue, circle, inner sep=0, minimum size=1mm] (1x1) at (0,0.4) {};
	\node[fill=none, circle, inner sep=0, minimum size=1mm] (1x2) at (0.4, 0.2) {};
	\node[fill=blue, circle, inner sep=0, minimum size=1mm] (1x3) at (0.8, 0.1) {};
	\node[fill=none, circle, inner sep=0, minimum size=1mm] (1x4) at (1.2, 0.2) {};
	\node[fill=blue, circle, inner sep=0, minimum size=1mm] (1x5) at (1.6, 0.7) {};
	\node[fill=blue, circle, inner sep=0, minimum size=1mm] (151) at (1.7, 0.75) {};
	\node[fill=blue, circle, inner sep=0, minimum size=1mm] (152) at (1.8, 0.8) {};
	\node[fill=none, circle, inner sep=0, minimum size=1mm] (1x6) at (2.0, 0.9) {};
	\node[fill=blue, circle, inner sep=0, minimum size=1mm] (1x7) at (2.4, 0.3) {};
	\node[fill=none, circle, inner sep=0, minimum size=1mm] (1x8) at (2.8, 0.1) {};
	\path[-, dash pattern={on 3pt off 3pt}, line width=0.2pt] (1x1.center) edge[out=0, in=180] (1x2.center);
	\path[-, dash pattern={on 3pt off 3pt}, line width=0.2pt] (1x2.center) edge[out=0. in=180] (1x3.center);
	\path[-, dash pattern={on 3pt off 3pt}, line width=0.2pt] (1x3.center) edge[out=0. in=180] (1x4.center);
	\path[-, dash pattern={on 3pt off 3pt}, line width=0.2pt] (1x4.center) edge[out=0, in=180] (1x5.center);
	\path[-, dash pattern={on 3pt off 3pt}, line width=0.2pt] (1x5.center) edge[out=0, in=180] (1x6.center);
	\path[-, dash pattern={on 3pt off 3pt}, line width=0.2pt] (1x6.center) edge[out=0, in=180] (1x7.center);
	\path[-, dash pattern={on 3pt off 3pt}, line width=0.2pt] (1x7.center) edge[out=0, in=180] (1x8.center);
	\path[->, very thick] (origin.center) edge (x.center);
	\path[->, very thick] (origin.center) edge node[below] [yshift=-1mm]{$t$} (x.center);
	\path[->, very thick] (origin.center) edge node[left, rotate=90, xshift=6mm, yshift=3mm] {Chan. 1} (y.center);
\end{scope}
	\draw[->] (1x1) -- (atn11.south);
	\draw[->, dotted] (1x3) -- (atn11.south);
	\draw[->, dotted] (1x5) -- (atn11.south);
	\draw[->, dotted] (151) -- (atn11.south);
	\draw[->, dotted] (151) -- (atn11.south);
	\draw[->] (1x7) -- (atn11.south);

	\draw[->] (1x1) -- (atn21.south);
	\draw[->, dotted] (1x3) -- (atn21.south);
	\draw[->, dotted] (1x5) -- (atn21.south);
	\draw[->, dotted] (151) -- (atn21.south);
	\draw[->, dotted] (151) -- (atn21.south);
	\draw[->] (1x7) -- (atn21.south);
\end{tikzpicture}

%% file: content/05_Toyexp.tex
\section{Modelling Goodwin Oscillators with FLD-L}\label{goodwin}
FLD operates on the assumption that complex functions can be modeled by combining multiple simple curves with a deep neural network. 
To investigate FLD-L's ability to learn non-linear dynamics, we conduct an
experiment with time series generated by the Goodwin oscillator
model~\cite{goodwin1965oscillatory}, which describes negative feedback interactions of cells at the molecular level. 

\begin{figure}
  \begin{minipage}{0.64\textwidth}
      \centering
      \begin{subfigure}{\linewidth}
          \includegraphics[width=\linewidth]{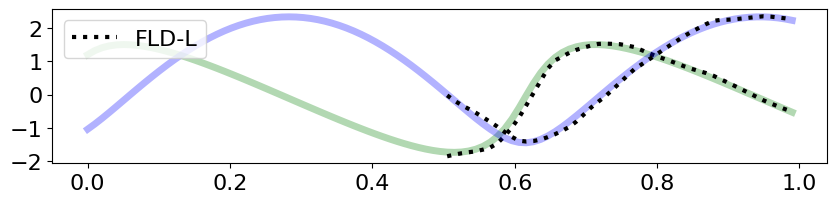}
          \caption{Ground truth and prediction}
          \label{fig:preds}
      \end{subfigure}
  \end{minipage}
  \hfill
  \begin{minipage}{0.35\textwidth}
      \centering
      \begin{subfigure}{\linewidth}
          \includegraphics[width=\linewidth]{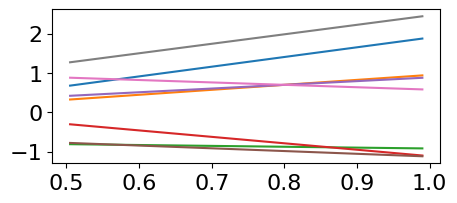}
          \caption{Hidden States}
          \label{fig:hidden}
      \end{subfigure}
  \end{minipage}
  \caption{Experiment on synthetic data created by the Goodwin oscillator model. 
  We show FLD-L's forecast (left) and the inferred hidden states (right).}
  \label{fig:goodwin}
\end{figure}

For our experiments we use the
implementation that was published in CellML~\cite{lloyd2008cellml}. The dataset
samples have two input channels and were generated by varying the constants and initial states of the
Goodwin oscillator. \Cref{fig:preds} shows a sample generated by the oscillator and FLD-L's prediction. 
Furthermore, we plot the hidden states that the trained FLD-L model inferred for that sample in \Cref{fig:hidden}.
The experiment on the synthetic Goodwin dataset demonstrates that FLD is capable to precisely infer a non-linear time series, although the 
hidden states develop linearly over time.

%% file: content/06_a_Benchdetails.tex
\section{Benchmark Experiments}
We provide details about the tasks, datasets, and models that were used in our experiments. To 
ensure a fair comparison with previous work, we utilize established benchmark datasets and 
protocols.
\subsection{Datasets}
Following the IMTS forecasting literature~\cite{de2019gru,bilovs2021neural,grafiti,scholz2023latent}, we conduct experiments on four different datasets USHCN, Physionet-2012, MIMIC-III and MIMIC-IV.
\begin{table}
	\centering
	\caption{Statistics of data sets used for experiments. \emph{Max. Len.} refers to the maximum sequence length among samples. \emph{Max. Obs.} refers to the maximum number of non-missing observations among samples. \emph{Sparsity} refers to the percentage of missing values over all samples.}
	\begin{tabular}{l|lllll}
		\hline
		Name                    & \#Sampl. & \#Chann. & Max. Len & Max. Obs & Spars. \\ \hline
		\textbf{USHCN}          & 1.114    & 5        & 370      & 398      & 78.0\% \\
		\textbf{Physionet-2012} & 11.981   & 37       & 48       & 606      & 80.4\% \\
		\textbf{MIMIC-III}      & 21.250   & 96       & 97       & 677      & 94.2\% \\
		\textbf{MIMIC-IV}       & 17.874   & 102      & 920      & 1642     & 97.8\% \\
		\hline
	\end{tabular}
	\label{tab:data}
\end{table}

\textbf{USHCN}~\cite{ushcn} contains measurements of 5 variables from 1280 weather stations in
the USA. Following the preprocessing proposed by DeBrouwer et al.~\cite{de2019gru}, most of the
150+ years of observation time is ignored and only measurements from 1996-2000
are used in the experiments. Furthermore, USHCN is made sparse artificially by
only keeping a randomly sampled 5\% of the measurements.

\textbf{Physionet-2012}~\cite{silva2012predicting} comprises a dataset
consisting of the medical records of 12,000 ICU-patients. During the initial 48
hours of admission, measurements of 37 vital signs were recorded. Following the
approach used in previous studies~\cite{de2019gru,bilovs2021neural,grafiti,scholz2023latent}, we pre-process the
dataset to create hourly observations, resulting in a maximum of 48 observations
in each series.

\textbf{MIMIC-III}~\cite{johnson2016mimic} is a widely utilized medical dataset
that provides valuable insights into the care of ICU patients. In order to
capture a diverse range of patient characteristics and medical conditions,
96 variables were meticulously observed and documented. To ensure
consistency, we followed the preprocessing steps outlined in previous studies
	[1, 3, 11]. Specifically, we round the recorded observations into 30-minute
intervals and only use observations from the 48 hours following the admission.
Patients who spend less than 48 hours in the ICU are disregarded.

\textbf{MIMIC-IV}~\cite{johnson2021mimic} represents an expansion and
improvement over MIMIC-III, offering an updated and enriched dataset that
enables more comprehensive exploration and analysis. It incorporates new data
sources and additional patient records, providing an enhanced foundation for
researchers to delve into temporal patterns, forecast future medical events, and
gain valuable insights into critical care management. Strictly following
\cite{bilovs2021neural,grafiti}, we preprocess MIMIC-IV similar to MIMIC-III,
but round observations into 1-minute intervals.

\subsection{Competing Models}
We compare FLD models against members of the neural ODE family: \textbf{GRU-ODE-Bayes}~\cite{de2019gru}, 
\textbf{Neural Flows}~\cite{bilovs2021neural}, \textbf{LinODENet}~\cite{scholz2023latent}, \textbf{CRU}~\cite{schirmer2022cru}. Besides the ODE-based models, we also compare our results to \textbf{GraFITi}~\cite{grafiti}, the state-of-the-art in IMTS forecasting. 

\textbf{mTAN}~\cite{shukla2021mTAN} was not introduced as a forecasting model, but we still selected the model as one of our competitors, because the FLD-Encoder is related to the mTAN encoder. The model is trained using the training routine that was originally proposed for interpolation purposes. In the experimental results of this work, mTAN refers to the mTAND-Full architecture, as described by Shukla et al.~\cite{shukla2021mTAN}.

\begin{table}[]
	\centering
	\caption{FLD's hyperparameter search space for the benchmark experiments}
	\begin{tabular}{p{6cm}l}
		\hline
		\textbf{Hyperparameter} & \textbf{Search Space} \\
		\hline
		Hidden Dimension        & \{32,128,256,512\}  \\
		Attention Heads         & \{4,8\}             \\
		Decoder Depth           & \{2,4\}             \\
		Embedding Size per Attention Head & \{2,4,8\}   \\
		\hline
	\end{tabular}
	\label{tab:hyps}
\end{table}

\subsection{Task Protocol}\label{sec:randomsearch}
We adopted the experimental protocol as published by Yalavarthi et al.~\cite{grafiti}.
Our experiments on IMTS forecasting involve varying the observation range and
forecasting horizon across multiple tasks for each dataset to assess different
model capabilities. The widely used 75\%-3 task requires
models to predict the next three time steps after observing 75\% of the time
series, equating to 36 hours for healthcare datasets and the first three years
for the USHCN dataset. To challenge models with a longer forecasting horizon, we
also undertake the 50\%-50\% task, where models predict the second half of an
IMTS using the first half as observations, meaning 24 hours of prediction for
medical datasets and 2 years for the USHCN dataset. Additionally, we also evaluate the models 
on the 75\%-25\% to add a task in between the two previous tasks. Here, models see the observations from the initial 36h / 3 years and forecast the remaining 12h / 1 year.
For hyperparameter search and early
stopping we take a validation set, consisting of 20\% of the available data.
Furthermore, we set aside another 10\%
of the data as unseen for the final evaluation (Test Data). 
We applied 5-fold cross-validation. Each fold reserves different subsets for validation and testing.

The implementation of the experiments are mainly based on the TSDM package
provided by Scholz et al.~\cite{scholz2023latent}. We run our experiments on Nvidia 2080TI GPU with 12GB.

\subsection{Hyperparameters}

Regarding hyperparameter optimization for competing models, we use the same hyperparameter search spaces and optimization protocol as introduced by Yalavarthi et al.~\cite{grafiti}. For each task, we randomly sample a maximum of 10 sets of hyperparameters and fully train
models with the respective configurations on one fold. We select the model with
the lowest MSE on the validation data of that fold and then train it on each of
the 5 folds to compute the mean and standard deviation of the test loss. The search space for the FLD models is described in \cref{tab:hyps}. While we vary the number of hidden layers in the decoder networks, we fix the width of each layer at the dimension of the hidden states $z$. 

For all models we use Adam optimizer~\cite{kingma2014adam}. For our models we use an initial learning rate of 0.0001.
Furthermore, we add an L2-regularization of weight 0.001.

%% file: content/06_b_benchresults.tex
\begin{table}[]
    \small
    \caption{Test MSE for forecasting next three time steps after 75\% observation time. 
    OOM refers to out of memory. $\dagger$: results reported by Yalavarthi et al.~\cite{grafiti}. 
    We highlight the best model in \textbf{bold} and the second best in \textit{italics}}  
    \centering
    \begin{tabular}{l|cccc}
        \hline
        & \textbf{USHCN} & \textbf{Physionet-12} & \textbf{MIMIC-III} & \textbf{MIMIC-IV} \\ 
        \hline
        \textbf{GraFITi}$^\dagger$ & 0.272 $\pm $ 0.047 & \textbf{0.286 $\pm$ 0.001} & 
        \textbf{0.396 $ \pm$ 0.030} & \textbf{0.225 $\pm$ 0.001}\\
        \textbf{mTAN}$^\dagger$   & $0.300 \pm 0.038$ & $0.315 \pm 0.002$ & $0.540 \pm 0.036$ & OOM \\
        \hdashline
        \textbf{GRU-ODE}$^\dagger$& $0.401 \pm 0.089$ & $0.329 \pm 0.004$ & $0.476 \pm 0.043$ & $0.360 \pm 0.001$ \\
        \textbf{Neural Flow}$^\dagger$& $0.414 \pm 0.102$ & $0.326 \pm 0.004$ & $0.477 \pm 0.041$ & $0.354 \pm 0.001$ \\
        \textbf{LinODE}$^\dagger$ & 0.300 $\pm$ 0.060 & 0.299 $\pm$ 0.001 & $0.446 \pm 0.033$ & $\mathit{0.272 \pm 0.002}$ \\
        \textbf{CRU}$^\dagger$    & $0.290 \pm 0.060$ & $0.379 \pm 0.003$ & $0.592 \pm 0.049$ &  OOM \\
        \hdashline
        \textbf{FLD-L} & $\mathit{0.262 \pm 0.040}$ & $\mathit{0.297 \pm 0.000}$ & $\mathit{0.444 \pm 0.027}$ & $0.274 \pm 0.000$ \\
        \textbf{FLD-Q} & $\mathbf{0.258 \pm 0.043}$ & $0.301 \pm 0.000$ & $0.451 \pm 0.024$ & $0.280 \pm 0.000$ \\
        \textbf{FLD-S} & $0.282 \pm 0.030$ & $0.307 \pm 0.000$ & $0.450 \pm 0.029$ & $0.313 \pm 0.002$ \\
        \hline
    \end{tabular}
    \label{tab:comp}
\end{table}

\subsection{Results}
We compare the forecasting accuracy of FLD-L, FLD-Q and FLD-S with that of the
competition by conducting experiments using various observation times and
forecasting horizons. Since we follow the experimental protocols from \cite{grafiti}, we
report their results whenever it is possible and run those experiments that have not been 
conducted yet.

\begin{table}[]
    \small
    \caption{Test MSE for forecasting next 25\% after 75\% observation time. OOM refers to out of memory. $\dagger$: results reported by Yalavarthi et al.~\cite{grafiti}. We highlight the best model in \textbf{bold} and the second best in \textit{italics}}   
    \centering
    \begin{tabular}{l|cccc}
        \hline
        & \textbf{USHCN} & \textbf{Physionet-12} & \textbf{MIMIC-III} & \textbf{MIMIC-IV} \\ 
        \hline
        \textbf{GraFITi} & $\mathbf{0.499 \pm 0.152} $& \emph{0.365 $\pm$ 0.001$^\dagger$} &$\mathbf{ 0.438  \pm 0.014^\dagger}$ &$\mathbf{0.285 \pm
         0.002^\dagger}$\\
        \textbf{mTAN} & 0.579 $\pm$ 0.182 & 0.514 $\pm$ 0.017 & 0.985 $\pm$ 0.055 & OOM \\
        \hdashline
        \textbf{GRU-ODE}$^\dagger$& 0.914 $\pm$ 0.343 & $0.432 \pm 0.003 ^\dagger$ & $0.591 \pm 0.018 ^\dagger$  & $0.366 \pm 0.154 ^\dagger$ \\
        \textbf{Neural Flow}& 1.019 $\pm$ 0.338 & $0.431 \pm 0.001 ^\dagger$ & $0.588 \pm 0.014 ^\dagger$ & $0.465 \pm 0.003 ^\dagger$ \\
        \textbf{LinODEnet} & 0.923 $\pm$ 0.877 & $0.373 \pm 0.001 ^\dagger$ & \emph{$0.477 \pm 0.021 ^\dagger$} & $0.335 \pm 0.002 ^\dagger$ \\
        \textbf{CRU} & 0.549 $\pm$ 0.238 & 0.435 $\pm$ 0.001$^\dagger$ & 0.575 $\pm$ 0.020$^\dagger$ &  OOM \\
        \hdashline
        \textbf{FLD-L} & $0.645 \pm 0.150$ & $\mathbf{0.360 \pm 0.001}$ & $\mathit{0.552 \pm 0.032}$ & $\mathit{0.321 \pm 0.000}$ \\
        \textbf{FLD-Q} & $0.601 \pm 0.097$ & $0.366 \pm 0.000$ & $0.559 \pm 0.028$ & $0.336 \pm 0.000$ \\
        \textbf{FLD-S} & $\mathit{0.526 \pm 0.205}$& $0.366 \pm 0.000$ & $0.558 \pm 0.033$ & $0.347 \pm 0.001$ \\
        \hline
    \end{tabular}
    \label{tab:comp7525}
\end{table}

\Cref{tab:comp}, \Cref{tab:comp7525} and \Cref{tab:comp5050} show the Test MSEs for each model on the 
75\%-3, 75\%-25\% and 50\%-50\% task respectively.
Based on our results we do not observe an FLD variant which consistently outperforms the other 
two members of the model family. For most datasets, FLD-L shows to be the best fit for the short and medium forecasting range, 
while FLD-S has the best accuracy on two datasets for the  50\%-50\% task among FLD variants. 
FLD-Q makes the best predictions on USHCN for
the 75\%-3 task, where it even surpasses the state-of-the art model GraFITi~\cite{grafiti}. However, USHCN carries large 
standard deviations across all models and task, especially for the longer forecasting ranges. 
Consequently, findings on this dataset are less conclusive. 
GraFITi reports superior forecasting accuracy's on 10 out 12 dataset/task combinations, but on 
Physionet-2012 and the 75\%-25\% task FLD-L improves on GraFITi making it the state-of-the art in this 
part of the evaluation. 
When we compare FLD's performance to the ODE-based models, we observe that the most accurate FLD variant 
outperforms the best ODE-based models in 7 out of 12 cases.
In particular, LinODENet outperforms, FLD on all datasets for the 50\%-50\% task.

\begin{table}[]
\small
\caption{Test MSE for forecasting next 50\% after 50\% observation time. OOM refers to out of memory. $\dagger$: results reported by
Yalavarthi et al.~\cite{grafiti}. We highlight the best model in \textbf{bold} and the second best in \textit{italics}}   
\centering
\begin{tabular}{l|cccc}
    \hline
    & \textbf{USHCN} & \textbf{Physionet-12} & \textbf{MIMIC-III} & \textbf{MIMIC-IV} \\ 
    \hline
    \textbf{GraFITi} & \textbf{0.623 $\pm$ 0.153} & $\mathbf{0.401 \pm 0.001^\dagger}$ & $\mathbf{0.491 \pm 0.014^\dagger}$ & 
    $\mathbf{0.285 \pm 0.002^\dagger}$ \\
    \textbf{mTAN} & \textit{0.721 $\pm$ 0.198} & 0.632 $\pm$ 0.023 & 1.016 $\pm$ 0.084 & OOM \\

    \hdashline
    \textbf{GRU-ODE}$^\dagger$& 1.019 $\pm$ 0.342 & $0.505 \pm 0.001^\dagger$ & $0.653 \pm 0.023^\dagger$  & $0.439 \pm 0.003^\dagger$ \\
    \textbf{Neural Flow}& 1.019 $\pm$ 0.338 & $0.506 \pm 0.002^\dagger$ & $0.651 \pm 0.017^\dagger$ & $0.465 \pm 0.003^\dagger$ \\
    \textbf{LinODEnet} & 0.724 $\pm$ 0.185 & $\mathit{0.411 \pm 0.001^\dagger}$ & $\mathit{0.531 \pm 0.022^\dagger}$ & $\mathit{0.336 \pm 0.002^\dagger}$ \\
    \textbf{CRU} & 0.729 $\pm$ 0.185 & 0.467 $\pm$ 0.002$^\dagger$ & 0.619 $\pm$ 0.028$^\dagger$ &  OOM \\
    \hdashline
    \textbf{FLD-L} & $0.874 \pm 0.212$ & $0.415 \pm 0.000$ & $0.545 \pm 0.026$ & $0.346 \pm 0.001$ \\
    \textbf{FLD-Q} & $0.888 \pm 0.236$ & $0.424 \pm 0.000$ & $0.554 \pm 0.025$ & $0.358 \pm 0.000$ \\
    \textbf{FLD-S} & $1.141 \pm 1.163$ & $0.414 \pm 0.000$ & $0.536 \pm 0.023$ & $0.359 \pm 0.001$ \\
    \hline
\end{tabular}
\label{tab:comp5050}
\end{table}

%% file: content/06_c_efficency.tex
\section{Efficiency}

We evaluate FLD's efficiency with respect to
inference time. For that experiment each benchmark model is trained
on the 50\%-50\% task of the Physionet-2012, MIMIC-III, MIMIC-IV, and USHCN
datasets. 

Efficiency comparison of machine learning models is a complex task, since different 
hyperparameter
configurations may introduce a trade-off between number of parameters and prediction accuracy.
Scholars typically compare the inference time of hyperparameter sets that were trained to 
optimize the training objective, in our case forecasting accuracy. However, we argue that 
this strategy is not necessarily fair to all models, since it ignores the trade-off between 
efficiency and accuracy. For example, there might exist a hyperparameter configuration that is 
barely suboptimal with regard to accuracy, but excels in terms of inference time. 
Consequently, we compare our model to the fastest hyperparameter configuration 
from each architecture's search space. This will provide a lower bound of inference time for
the competing models and is only unfair to FLD. 

\begin{table}
    \small
    \caption{Comparison of inference time in seconds on the 50\%-50\% task. OOM indicates a memory error.}
    \centering
    \begin{tabular}{l|llll}
    \hline
     &\textbf{USHCN} & \textbf{Physionet-12} & \textbf{MIMIC-III} & \textbf{MIMIC-IV}  \\ 
    \hline
    \textbf{GraFITi}  &  0.176  & 2.775  & 3.640    & 6.719\\   
    \textbf{mTAN} & 0.062  & 0.776  & 1.068    & 3.494 \\
    \hdashline
    \textbf{GRU-ODE} & 5.378  & 38.118 & 46.272   & 154.543  \\
    \textbf{Neural Flow}&  1.630  & 2.835    & 6.428 & 44.187  \\
    \textbf{f-CRU}   & 1.657  & 4.578  & 9.281    & OOM  \\   
    \textbf{LinODE}  & 2.852  & 6.294  & 13.776   & 95.050 \\ 
    \hdashline
    \textbf{FLD-L} & $0.018$ & $0.237$ & $0.394$ & $2.141$ \\
    \textbf{FLD-Q} & $0.020$ & $0.243$ & $0.431$ & $2.380$ \\
    \textbf{FLD-S} & $0.021$ & $0.245$ & $0.435$ & $2.740$ \\
    \hline
    \end{tabular}
    \label{tab:inf}
\end{table}

We assume that for each model the smallest hyperparameter instances provide the fastest inference. For example, with Neural
Flows~\cite{bilovs2021neural}, we use only 1 flow layer, as employing
multiple flow layers leads to slower inference and training.
We opt for the \emph{euler} solver instead of the \emph{dopri5} solver due to its significantly faster inference, to infer the hidden state of GRU-ODE-Bayes~\cite{de2019gru}. 
Additionally, we use the \emph{fast} variant of CRU (f-CRU), that was 
introduced by Schirmer et al.~\cite{schirmer2022cru}. For FLD-L, we use the hyperparameter set that has been tuned on validation loss for
each task because we found a negligible change in computational speed for
different hyperparameters.
Table~\ref{tab:inf} reports the inference time of
each model on various datasets. More specific, we refer to the wall-clock time 
to predict the complete test data of each dataset, with a batch size of 64.
We observe that FLD-L infers predictions
significantly faster than the competing models. Since mTAN is the second-fastest model, we 
conclude that FLD's speed is related to the performant attention-based encoder, since it is closely 
related with the mTAN encoder. FLD's inference 
with parameterized curves results in fewer operations and a significant gain in computational speed. Furthermore, keep in mind that mTAN's inference time increases drastically if we add more reference points and parameters. 

\begin{figure}[t]
    \begin{minipage}{0.48\textwidth}
        \centering
        \begin{subfigure}{\linewidth}
            \includegraphics[width=\linewidth]{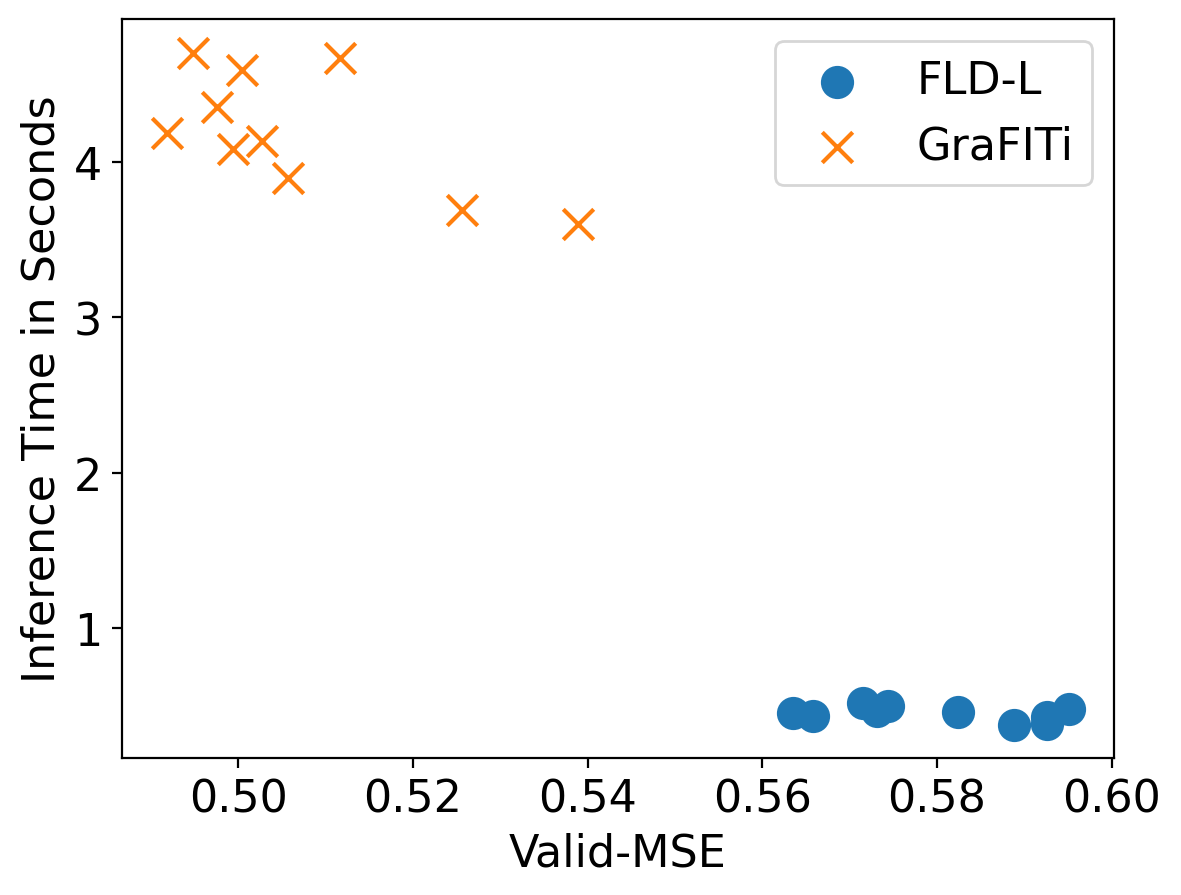}
            \caption{MIMIC-III}
            \label{fig:m3effi}
        \end{subfigure}
    \end{minipage}
    \hfill
    \begin{minipage}{0.48\textwidth}
        \centering
        \begin{subfigure}{\linewidth}
            \includegraphics[width=\linewidth]{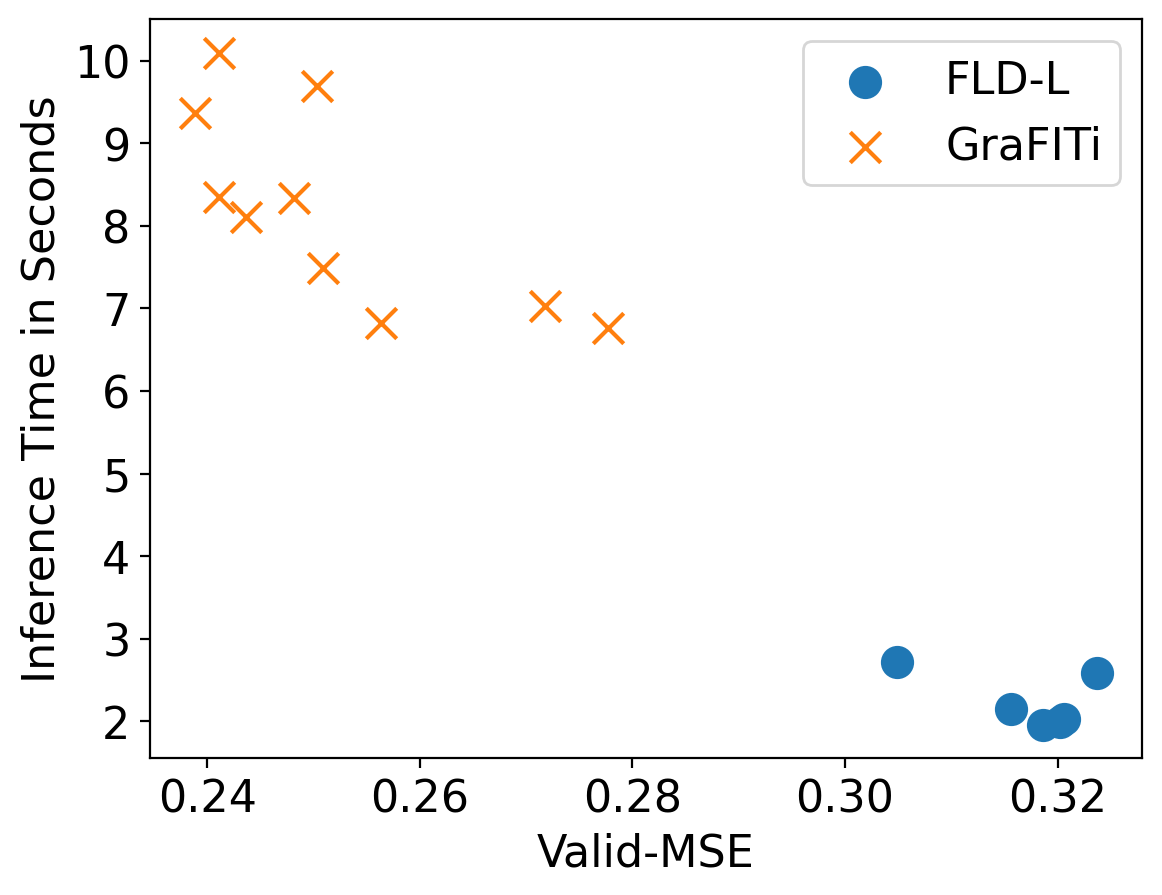}
            \caption{MIMIC-IV}
            \label{fig:m4effi}
        \end{subfigure}
    \end{minipage}
    \caption{Efficiency comparison of FLD-L and GraFITi. We plot the validation loss and inference
    time for 10 randomly sampled hyperparameter configurations for each GraFITi and FLD-L. 
    The plots refer to results on the 75\%-25\% task on MIMIC-III and MIMIC-IV.}
    \label{fig:effigraf}
\end{figure}

To gain more insight into the trade-off between inference time and forecasting accuracy, we conduct 
a more detailed efficiency comparison. In \Cref{fig:effigraf}
we plot validation MSE and inference time of 10 randomly sampled hyperparameter configurations of 
FLD-L and GraFITi on the two largest datasets MIMIC-III and MIMIC-IV. The plot shows that all 
versions of FLD-L were significantly faster than GraFITi. However, they were also constantly inferior with 
respect to forecasting accuracy.

%% file: content/07_Concl.tex
\section{Conclusion and Future Work}
In this work, we introduced a novel approach to forecast irregularly sampled multivariate
time series (IMTS). In particular, we proposed Functional Latent Dynamics (FLD),
a model family that models the hidden state of an IMTS with a continuous curve
function. This serves as an efficient and accurate alternative to ODE-based
models, which have to solve complex differential equations. To be more specific,
we outperform all ODE-based models in task with a short and medium forecasting range.
Additionally, we surpass the IMTS forecasting state-of-the-art model 
GraFITi~\cite{grafiti} on 2 of 12 evaluation tasks. Our models have magnitudes 
faster inference speed when compared to ODE approaches, and multitudes faster inference 
speed than GraFITi.

Our FLD-Encoder can elegantly handle missing observations
in order to compute the coefficients of the curve functions.
Even if the hidden states are linear, FLD can learn
to forecast non-linear functions since non-linearity is induced by its decoder.
We demonstrate that hidden states that follow linear curve functions 
are expressive enough to imitate Goodwin oscillators.

In the future, we will tackle the problem of combining different forms of
curve functions like sine and linear curves. Here, the distant vision is
to \emph{learn} which kind of curves are appropriate for a specific time-series
dataset. As our results indicate that FLD is a performant approach for
time-series forecasting, it is promising to transfer it to probabilistic
forecasting settings. Here, it is crucial to derive possibilities for FLD to
output distributions instead of point predictions. To achieve this, FLD can, for
example, be used as an encoder for a conditioning input for a normalizing flow.

